\documentclass[letterpaper, 10 pt, conference]{ieeeconf}  
\pdfoutput=1
\IEEEoverridecommandlockouts                              
\makeatletter
\let\NAT@parse\undefined
\makeatother

\usepackage{graphics} 
\usepackage{epsfig} 
\usepackage{amsmath} 
\usepackage{amssymb}  

\usepackage{mathrsfs}
\usepackage[colorlinks=true, citecolor=red, linkcolor=red, urlcolor=black, bookmarks=true, bookmarksnumbered=true, pdftex, plainpages=false, pdfpagelabels,backref=true]{hyperref}

\title{\LARGE \bf
Receding horizon estimation and control with structured noise blocking for mobile robot slip compensation
}

\author{Nathan Wallace, He Kong, Andrew Hill and Salah Sukkarieh$^{1}$
\thanks{$^{1}$The authors are with the Australian Centre for Field Robotics (ACFR), The University of Sydney, NSW 2006, Australia. Corresponding authors:
        {\tt\small \{n.wallace,h.kong\}@acfr.usyd.edu.au}}%
}

\begin{document}

\maketitle
\thispagestyle{empty}
\pagestyle{empty}

\begin{abstract}

The control of field robots in varying and uncertain terrain conditions presents a challenge for autonomous navigation. Online estimation of the wheel-terrain slip characteristics is essential for generating the accurate control predictions necessary for tracking trajectories in off-road environments. Receding horizon estimation (RHE) provides a powerful framework for constrained estimation, and when combined with receding horizon control (RHC), yields an adaptive optimisation-based control method. Presently, such methods assume slip to be constant over the estimation horizon, while our proposed structured blocking approach relaxes this assumption, resulting in improved state and parameter estimation. We demonstrate and compare the performance of this method in simulation, and propose an overlapping-block strategy to ameliorate some of the limitations encountered in applying noise-blocking in a receding horizon estimation and control (RHEC) context.

\end{abstract}

\section{INTRODUCTION}

In recent years the agriculture industry has increasingly adopted the use of autonomous vehicles and machinery in an attempt to handle large workloads under stringent time pressures, and to offset the high expenses and unreliable availability of manpower. The large scale of many agricultural operations necessitates traversing long distances over often rough and varying terrain conditions in an expedient manner in order to meet time constraints.

Accurate navigation in these difficult off-road conditions presents one of the major challenges in development and deployment of autonomous agents for applications such as farming and environmental monitoring. The presence of unknown and spatio-temporally varying traction conditions can degrade the performance of traditional control approaches (eg. PID control) \cite{Fukao2000} which do not account for the impact of induced wheel slippage on vehicle handling. It is therefore necessary to estimate the traction parameters online, so that the control algorithm can compensate for wheel slippage and maintain good tracking performance. This is important in situations where the robot is operating near crops or livestock, to ensure sufficient clearance is maintained.

In this paper, we implement a non-linear receding horizon estimation and control (RHEC) approach for tracking trajectories in unknown and variable slip environments. Building upon insights in our prior work \cite{He2018}, we investigate the impact of enforcing a noise-blocking structure on the estimated slip parameter sequence in terms of the estimation and tracking performance within the RHEC framework. We then propose an extension to this strategy that permits block overlap at the end of the estimation horizon, reducing the periodic variation in parameter estimation quality that emerges from the enforced blocking structure.

We compare the results of our modified RHEC method with the work of Kraus et. al. \cite{Kraus2013}, which adopts an RHEC approach with the slip parameters fixed to a single value over the estimation horizon - henceforth termed full-horizon blocking (FHB). We show that the proposed structured blocking method exhibits both improved state and parameter estimation performance across a variety of different speeds (1-10m/s). The impact of block size and overlap extent on performance is also evaluated.

\section{RELATED WORK}

Achieving robust control of wheeled mobile robot (WMR) systems in the presence of modelling errors and external disturbances is a long-standing challenge in robotics. Early works in this area study control in the presence of bounded additive disturbances \cite{Dixon2000}-\cite{Orlando2002}, and while these techniques can be specialised for scenarios with slip, deriving the bounds for the uncertainties presents a challenge.

The recent literature on WMR control with slip can be classified into three main slip modelling approaches; bounded uncertainty models, extended kinematic models (KM) which account for slip effects, and learnt slip models.

\subsection{Slip modelling using bounded uncertainties}
Kinematic representations relating perturbations to vehicle slip are presented in \cite{Wang2008TRO} for four general WMR configurations, classifying slip perturbations as input-additive, input-multiplicative or matched/unmatched. These are utilised in \cite{Wang2010} to design path following controllers which, assuming no other disturbances, apply backstepping techniques and prove the ultimate boundedness of the tracking errors.

In \cite{Savkin2013}, the effects of wheel slip for a tractor are treated as bounded uncertainties, and a robust sliding-mode controller is designed to achieve stability with steering angle constraints. Similar ideas treating slips as bounded uncertainties have also been adopted in \cite{Ramon2015}-\cite{Alamo2011} using tube-based model predictive control techniques for vehicle speeds of $<$1 m/s.

\subsection{Slip modelling using extended kinematic models}

Slip estimation for a skid-steered mobile robot is investigated in \cite{Yi2009}, using a KM for analysis of the wheel-slip relationships with respect to the robot motion. Longitudinal slip is modelled as a multiplicative factor, and an extended Kalman filter (EKF) design incorporating kinematic constraints is used to improve estimation accuracy.

The impact of slip has also been modelled as a white-noise velocity perturbation \cite{Sukkarieh2001}, and an EKF with vehicle model constraints was used to improve the vehicle's inertial estimate in the absence of GPS data.

In \cite{Backman2012} a generalized bicycle KM with an additional constant multiplicative factor (within trigonometric functions) is introduced to model a tractor-trailer system with side slipping effects. An EKF is used to estimate the state and slip parameters which a non-linear RHC uses for path tracking, and this strategy achieved a lateral tracking error for the trailer of $<$10 cm and $<$15 cm on straight and gently curved paths respectively at 3.3 m/s.

A similar method for modelling slip was adopted in \cite{Kraus2013} for a tractor, augmenting the well-known bicycle KM with two multiplicative factors capturing longitudinal and side slip, which are assumed to be constant over the estimation horizon. State and parameter constraints, and a nonlinear measurement function, necessitated use of the more advanced RHE strategy, rather than an EKF. An RHC is proposed to control the wheel velocity and steering rate, and this RHEC framework is tested experimentally on a wet and bumpy grass field, with average deviations from the time-based reference trajectory of 0.26 m on slightly curved strips, and 0.6 m in headland turns at speeds of $\sim$2 m/s. A similar approach was applied for a tracked field robot in \cite{Kayacan2018}, using a unicycle KM augmented with two traction parameters representing longitudinal and lateral slip. Tracking error was shown to be $<$12 cm for speeds of $\sim$0.2 m/s.

Ideas akin to those in \cite{Kraus2013} have been adopted for a tractor-trailer system \cite{Kayacan2014}, using a generalised tricycle KM augmented with three multiplicative factors describing the longitudinal and side slips of the tractor and trailer. A centralised RHE strategy performs the estimation, and a distributed RHC framework is proposed and compared against a centralised RHC strategy. Tracking error was shown to be $<$10 cm, with the decentralised strategy yielding faster solutions.

In \cite{Thuilot2006} a tractor is modelled with an extended bicycle KM, with additive side slip effects on both wheels, while not considering longitudinal slip. This model is transformed into a chain form, and a non-linear adaptive control law embedded with an RHC algorithm to avoid path tracking overshoot is derived and shown experimentally to yield tracking errors below 15 cm, with wheel velocity controlled manually by a human driver.

It was observed in \cite{Thuilot2010} that the techniques in \cite{Thuilot2006} suffer from decreased performance as vehicle speed increases, motivating development of a mixed backstepping kinematic and dynamic observer to improve observation of the side slip effects. This new framework was reported to improve tracking performance at the maximal velocity of 5 m/s.

\subsection{Slip modelling using learnt models}

Learning-based methods aim to learn slip behaviours from prior experience and apply these learnt models predictively. Remote prediction using visual information \cite{Angelova2007}-\cite{Ostafew2016} can be used to learn a non-linear disturbance model which, in conjunction with a simple a priori vehicle model, has been shown to achieve high performance path tracking at speeds up to 2 m/s. However, the robustness of learning controllers remains a largely unanswered question, though some works have attempted to address this via checking against a nominal vehicle model \cite{Aswani2013}. Reinforcement learning methods have also been applied \cite{Abbeel2006}, although prohibitively large amounts of training data are necessary to sufficiently generalise the learnt model for operation.

\section{PRELIMINARIES}

\subsection{Notation}

Throughout this paper, $\left[a_1, \dots, a_n\right]$ denotes $\left[a_1^T, \dots, a_n^T\right]^T$, where $a_1, \dots, a_n$ are appropriately dimensioned scalars/vectors/matrices. The weighted Euclidean norm is denoted by $\begin{Vmatrix}a\end{Vmatrix}^2_{R} = a^T R a$. Subscripts $i,j,k$ denote the value of that variable at time $t_k$, ie $a_k = a(t_k)$. $\mathscr{I}^{j}_{i}$ denotes the set of integers between $i$ and $j$.

\begin{figure}[tpb]
	\vspace{0.2cm}
	\centering
	\includegraphics[scale=0.45]{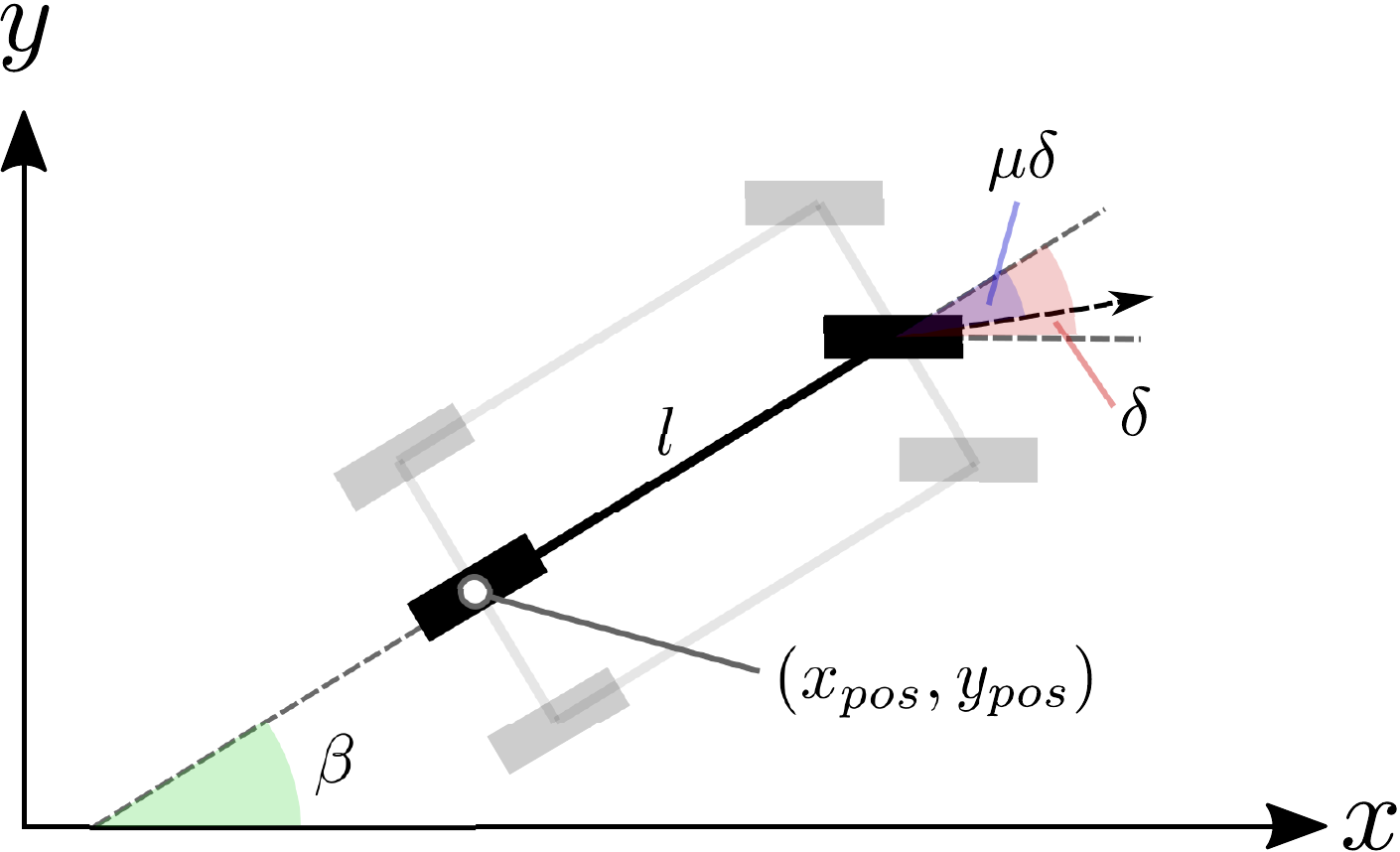}
	\caption{Schematic illustration of adapted bicycle model with side slip affecting the front wheel. The steering angle $\delta$ is shown in red, and the effective steering angle $\mu\delta$ is shown in blue.}
	\label{fig:bicycleModel}
	\vspace{-0.2cm}
\end{figure}

\subsection{System Model}

For simplicity, the vehicle model used in this study is identical to that used in \cite{Kraus2013}; a slight adaptation of the conventional bicycle KM, which is known to approximate 4-wheeled vehicle behaviour quite accurately \cite{LaValle2006}-\cite{Kong2015}. Slip is incorporated via two multiplicative parameters; one for longitudinal slip and another for side slip. 

The equations describing the motion are as follows:
\begin{align}
\begin{bmatrix}
\dot{x}_{pos} \\
\dot{y}_{pos} \\
\dot{\beta} \\
\dot{\delta}
\end{bmatrix}
= 
\begin{bmatrix}
\kappa u_{1} \cos{\beta} \\
\kappa u_{1} \sin{\beta} \\
\frac{\kappa u_{1}}{l} \tan\left(\mu\cdot\delta\right) \\
u_{2}
\end{bmatrix}, \label{eqn:sysModel}
\end{align}

\noindent
where $x_{pos}$ [m] and $y_{pos}$ [m] represent the position of the `rear wheel' - or centre of the rear axle in the case of a 4-wheeled vehicle - in global cartesian coordinates, $\beta$ [rad] is the yaw angle or heading, and $\delta$ [rad] is the steering angle; shown in Figure \ref{fig:bicycleModel}. The control variables are wheel speed with respect to the ground, $u_{1}$ [m/s], and the steering rate $u_{2}$ [rad/s]. The parameter $l$ [m] represents the wheel base.

The parameters $\kappa, \mu \in [0,1]$ represent the longitudinal and side slip respectively. $\kappa$ relates commanded wheel speed to actual ground speed, and $\mu$ results from inertia of the vehicle, which is approximately relative to the steering angle. The percentage slip experienced is therefore $1-\kappa$ and $1-\mu$.

\section{RECEDING HORIZON ESTIMATION AND CONTROL} \label{sec:RHEC}

\begin{figure}[tpb]
	\vspace{0.2cm}
	\centering
	\includegraphics[scale=0.45]{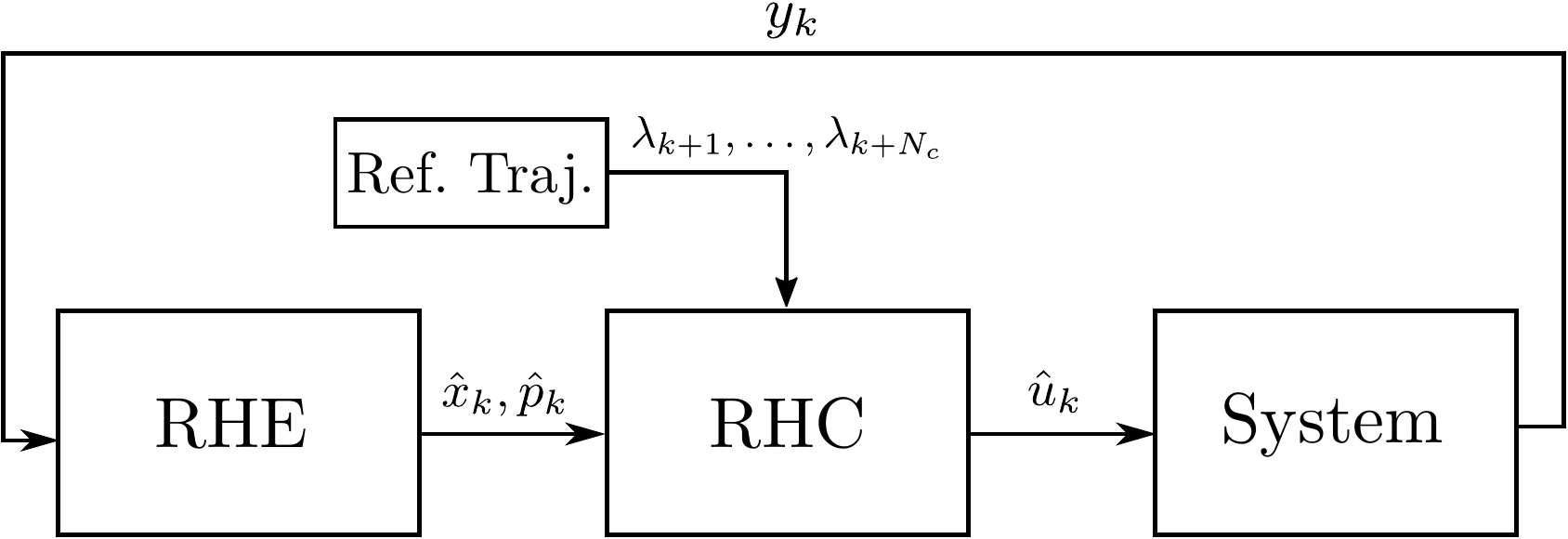}
	\caption{Block diagram of the RHEC framework.}
	\label{fig:RHECframework}
	\vspace{-0.2cm}
\end{figure}

A RHEC strategy is adopted in this work, the structure of which is outlined in Figure \ref{fig:RHECframework}. The non-linear RHEC problem is formulated as follows. Let $x_k \in \mathbb{R}^{n_x}$ represent the state of the system, $y_k \in \mathbb{R}^{n_y}$ an observation, $u_k \in \mathbb{R}^{n_u}$ the control actions and $p_k \in \mathbb{R}^{n_p}$ the system parameters we wish to estimate. The system is assumed to evolve in accordance to the given dynamic model $\dot{x}_k = f\left(x_k, u_k, p_k\right)$, and observations are taken in accordance with the measurement model function $y_k = h\left(x_k,u_k,p_k\right)$. Let variables $\left(x_k, y_k, u_k, p_k\right)$ refer to the \textit{real} process. These each have associated \textit{decision} and \textit{optimal decision} variables in the optimisation, which we denote $\left(\chi_k, \eta_k, \nu_k, \rho_k\right)$ and $\left(\hat{x}_k, \hat{y}_k, \hat{u}_k, \hat{p}_k\right)$ respectively.

At each sampling time interval $t_k$, let the real system state be $x_k$. A sensor measurement $y_k$ is taken, and the RHE then uses the past $N_e$ measurements $y_j,\ j \in \mathscr{I}^{k}_{k-N_e+1}$ to estimate $(x_k, p_k)$. This estimate $\left(\hat{x}_k,\ \hat{p}_k\right)$ is passed to the RHC module along with a time-based reference trajectory $\lambda_j = \begin{bmatrix}\lambda^x_j, \lambda^u_j\end{bmatrix}, \hspace{0.5em} \lambda^x_j \subseteq \mathbb{R}^{n_x}, \lambda^u_k \subseteq \mathbb{R}^{n_u}, j \in \mathscr{I}^{k+N_c}_{k+1}$. The RHC module then computes the optimal control actions over the horizon $N_c$, and the next action $\hat{u}_k$ is sent to the robot for execution. This process repeats at each sampling interval (spaced $\Delta t$ seconds apart) until the end of the reference trajectory is reached.

\subsection{Non-linear Receding Horizon Estimation} \label{sec:RHE}

Receding horizon estimation (RHE) is a powerful optimisation-based estimation technique that provides a systematic framework for handling constraints. In contrast to full information estimation (FIE), which utilises the entire history of available information, RHE uses only measurements taken within a finite time-frame, and captures the information of prior measurements in an arrival cost term.

Given $N_e$ past measurements $y_j,\ j \in \mathscr{I}_{k-N_e+1}^k$ and a measurement model function $h$, the constrained estimation problem to be solved at time $t_k$ is

\begin{align}
\min\limits_{\chi(\cdot),\nu(\cdot),\rho(\cdot)} && \Bigg{\lbrace}\Gamma_{k-N_e+1} + \sum_{i = k-N_e+1}^{k}\begin{Vmatrix}\eta_i-y_i\end{Vmatrix}^2_{R_k}\Bigg{\rbrace} \label{eqn:RHE}
\end{align}
\begin{align}
\begin{array}{l}s.t.\\ \\ \\ \\\end{array} && \left.\begin{array}{l}
\dot{\chi}_j = f\left(\chi_j,\nu_j,\rho_j\right)\\
\eta_j = h\left(\chi_j,\nu_j,\rho_j\right)\\
\chi_{min} \leq \chi_j \leq \chi_{max}\\
\rho_{min,j} \leq \rho_j \leq \rho_{max,j}
\end{array} \right\} & \text{ } j \in \mathscr{I}_{k-N_e+1}^k\nonumber
\end{align}
where 
\begin{align}
\Gamma_{k-N_e+1} = \begin{Vmatrix}\chi_{k-N_e+1} - \hat{x}_{k-N_e+1}\\\rho_{k-N_e+1} - \hat{p}_{k-N_e+1}\end{Vmatrix}^2_{\Pi^{-1}_{k-N_e+1}},
\end{align}
\noindent
is the arrival cost function, $\hat{x}_{k-N_e+1} = \hat{x}_{k-N_e+1|k-N_e}$ is the optimal state prediction at time $t_{k-N_e+1}$, $R_k$ is the symmetric positive semi-definite weighting matrix equal to the inverse of the measurement noise covariance matrix \cite{Ferreau2012}, and $\Pi^{-1}$ is the inverted Kalman covariance \cite{Rawlings2001}.

\subsection{Structured noise blocking in estimation} \label{sec:SB}

One of the main issues encountered with constrained estimation methods such as RHE and especially FIE, is that an optimisation problem must be solved at each sampling time. This burden can be alleviated via reformulation as a multi-parametric quadratic program \cite{Voelker2013}, or exploiting the optimisation structure \cite{Pannocchia2015}. 

A strategy called move blocking (MB) has seen prior application in RHC for a similar purpose \cite{Lee2015}. This strategy involves constraining groups of adjacent-in-time predicted inputs to have the same value, curtailing the number of degrees of freedom in the optimisation problem to reduce complexity at the expense of optimality \cite{Shekhar2015}. 

\begin{figure}[tpb]
	\vspace{0.2cm}
	\centering
	\includegraphics[scale=0.48]{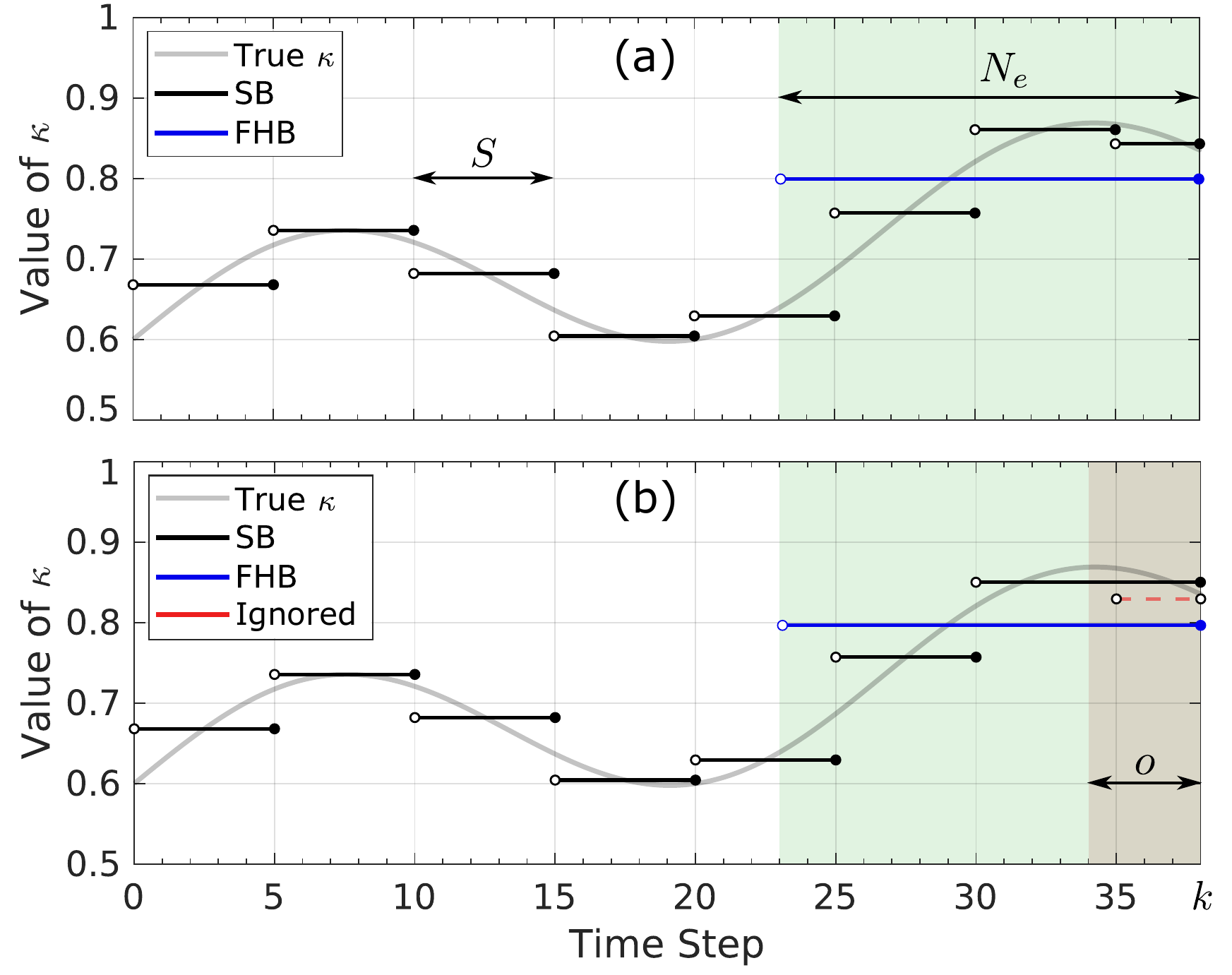}\vspace{-0.2cm}
	\caption{Illustration of the structured blocking (SB) concept. (a) The FHE MB structure is shown for a block size $S=5$; the RHE will observe only the last $N_e$ time steps. At time step $k = 38$, the last and earliest blocks in the horizon are partial ones. The equivalent FHB block is shown in blue. (b) The same scenario with an overlap $o=4$. In this case, the last block in the sequence (red) has size $3<o$, so it is suppressed, and the second last block is extended up to time step $k$. At time step $k+1$, the last block would have size $4 \nless o$, and would no longer be ignored.}
	\label{fig:noiseblocking}
	\vspace{-0.5cm}
\end{figure}

Recently, we have adopted a similar concept in FIE and RHE for estimating the process noise sequence (PNS) \cite{He2018}, and more importantly, we show that to ensure stability in RHE one has to enforce the same segment structure of the PNS enforced in MB FIE for the optimization steps within the receding horizon (i.e., steps between $T-N$ and $T-1$). The resulting MB RHE strategy becomes a dynamic estimator with a periodically varying computational complexity. An illustrative example of this concept, henceforth termed structured blocking (SB), is shown in Figure \ref{fig:noiseblocking} (a).

The contribution presented in this work is to extend the noise-blocking concept to parameter and state estimation, examining the strategy within the context of RHEC for slip parameter estimation.

To that end, we can consider the slip parameters $\kappa,\mu$ to be conceptually analogous to environmental noise. Thus, the estimator will choose values of these parameters such that the deviation between the estimated and measured state is minimised. However, the optimal choice of value for these parameters from an optimisation perspective is not necessarily reflective of the true value, as in the unblocked case it can result in overfitting to the measurements.

In the context of RHEC, we must be particularly attentive of this behaviour, as the controller relies on the estimator for the parameter values, and an incorrect estimate can result in a drastically sub-optimal control decision. A simple approach to resolving this issue was adopted in \cite{Kraus2013} and \cite{Kayacan2018} where the values of the slip parameters were assumed constant over the 3s estimation horizon. This produced good results at the speeds tested in those works, typically below $2.5$ m/s, however we conjecture that at higher vehicle speeds, where the rate of change in the parameter values is increased with respect to the vehicle motion, the state and parameter estimation accuracy of this method will be compromised.

By applying SB it is possible to attain better parameter estimates than the unblocked case. However, in enforcing the segment structure, we observed that the final block in the horizon will vary in size from $[1, S]$. When this final block is small, the parameter estimation runs into the same overfitting issue as in the unblocked case, albeit with a reduced quality state estimate due to the prior blocking as well.

This has lead us to propose a strategy to address this issue; virtual extension, or `overlapping', of blocks. Functionally, this means we both suppress the existence of the last block in the horizon unless its size is at least equal to an overlap parameter $o \in \mathbb{Z} \leq S$, and increase the extent of the prior block to include the overlap region $\mathscr{I}^{t_k}_{t_k-o\Delta t}$. An illustration of the SB with overlapping (SB-O) is shown in Figure \ref{fig:noiseblocking} (b). This strategy reduces the variability in the final block's parameter estimate, and thus provides a more reliable estimate to the RHC module.

\subsection{Non-linear Receding Horizon Control} \label{sec:RHC}

Receding horizon control (RHC) is an approach that seeks to predict the system behaviour over a finite time horizon via minimisation of a cost function composed of states, inputs and references, and similarly to RHE, the RHC framework also supports handling of state and input constraints. 

For RHC, at the current time $t_k$ we wish to predict the state-action-parameter sequence for the next $N_c$ time steps; thus the constrained optimisation problem to be solved is:

\begin{align}
\min\limits_{\chi(\cdot),\nu(\cdot)} && \Bigg{\lbrace}\sum_{i = k+1}^{k+N_c-1}\begin{Vmatrix}\lambda^x_i-\chi^\lambda_i\\\lambda^u_i-\nu^\lambda_i\end{Vmatrix}^2_{V_k}  + \Omega_{k+N_c}\Bigg{\rbrace}\label{eqn:RHC}
\end{align}
\begin{align}
\begin{array}{l}s.t.\\ \\ \\\end{array} && \left.\begin{array}{l}
\dot{\chi}_j = f\left(\chi_j,\nu_j,\rho_j\right)\\
\chi_{min} \leq \chi_j \leq \chi_{max}\\
\nu_{min} \leq \nu_j \leq \nu_{max}
\end{array} \right\} & \text{ } j \in \mathscr{I}_{k+1}^{k+N_c}\nonumber
\end{align}
where 
\begin{align}
\Omega_{k+N_c} = \begin{Vmatrix}\lambda_{k+N_c} - \chi^\lambda_{k+N_c}\end{Vmatrix}^2_{V_{N}},
\end{align}
\noindent
is the terminal cost function, $\chi^\lambda_k \in \mathbb{R}^{n_\lambda}$ denotes the subset of the state variable set $\chi_k$ with corresponding entries in $\lambda_k$, and $V_k, V_N$ are symmetric positive-semidefinite weighting matrices.

\section{Solution Methods}

As both RHE (\ref{eqn:RHE}) and RHC (\ref{eqn:RHC}) are non-linear least squares optimisation problems at heart, similar solution methods can be applied to both. Three popular approaches for such optimisation problems include simultaneous collocation, single-shooting, and multiple-shooting methods.

In this work, the generalised Gauss-Newton multiple-shooting method derived from the classical Newton method is used. It was developed specifically for solving least-squares optimisation problems quickly, in that it does not require calculating the second derivatives, thereby avoiding those difficult computations. However, it is challenging to determine the number of iterations necessary to reach a suitably accurate solution. The real-time iteration scheme proposed in \cite{Diehl2002} solves this issue by restricting the number of iterations to $1$,  and the initial value of each optimisation problem takes on the value of the previous one intelligently, improving convergence and reducing computation cost.

To solve the constrained nonlinear optimisation problems involved in RHEC, the \textit{ACADO} code-generation tool was used; an open-source software package which allows the user to export customised real-time RHE and RHC algorithms as efficient, self-contained C-code modules \cite{Houska2011}. The generated dense quadratic sub-problems are then solved using the \textit{qpOASES} online quadratic program solver \cite{Ferreau2014}.

\section{EXPERIMENTAL SETUP} \label{sec:exptSetup}

The goal of this experiment is to investigate how SB impacts the estimation and control performance of a robot tracking a pre-defined space-based trajectory through varying slip conditions. The choice of using a space-based rather than time-based trajectory was made due to the fact that the unknown slip conditions can have a significant impact on the robot speed, and the time deviations arising from these uncertainties can challenge the tracking of a time-based reference trajectory. With a space-based trajectory, the robot is not forced to be at any given point on the trajectory at a particular time-step.

\begin{figure}[tpb]
	\vspace{0.2cm}
	\centering
	\includegraphics[scale=0.42]{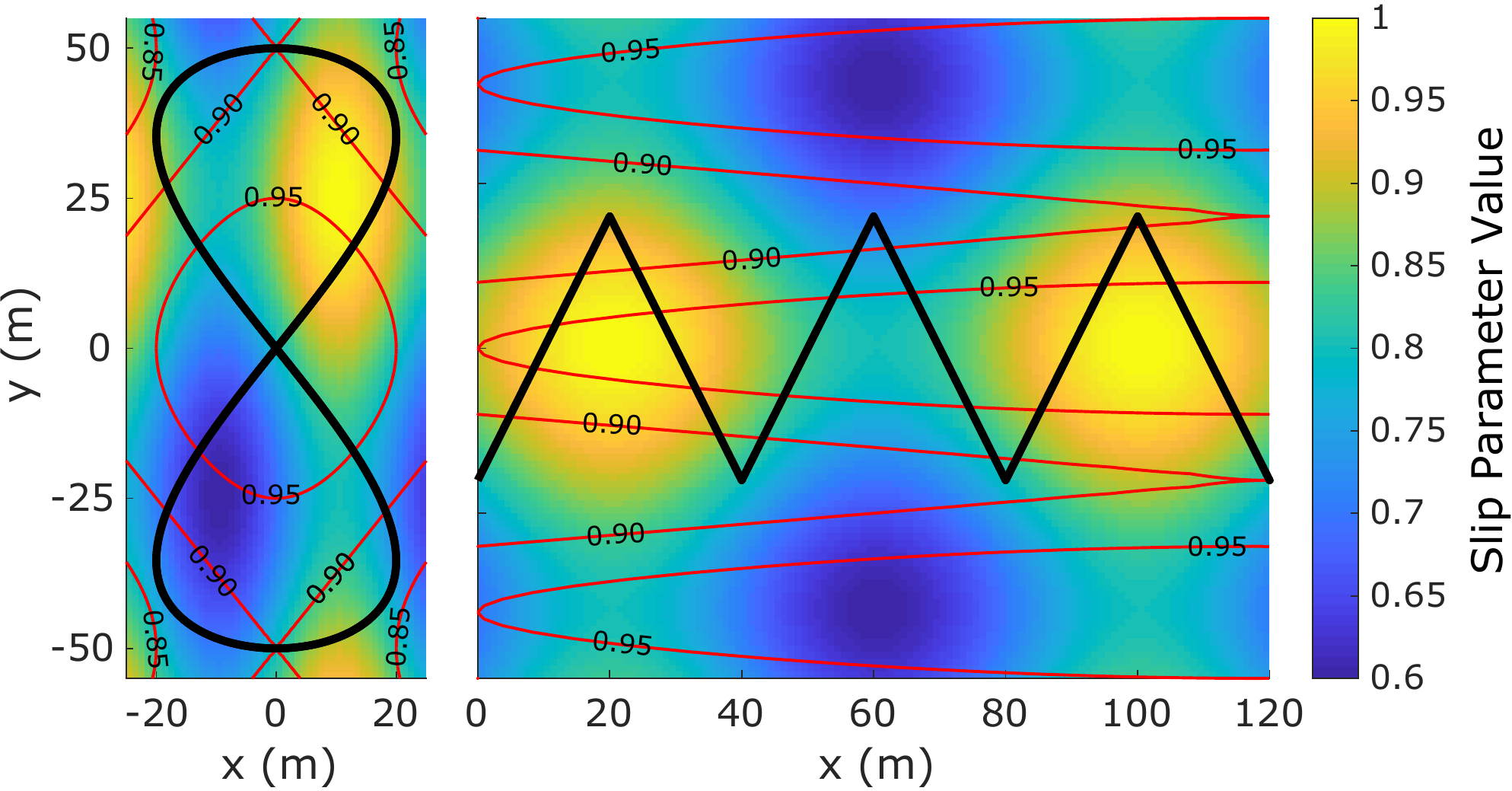}
	\vspace{-0.2cm}
	\caption{Reference trajectories plotted over spatial map of true parameter values; $\kappa$ in base map, $mu$ with contours. Figure-8 path is navigated twice, and first half of zig-zag path is shown. Length $\sim 550$m for both.}
	\label{fig:refTraj}
	\vspace{-0.5cm}
\end{figure}

Two different reference paths were used in the comparisons in this experiment; a figure-8 pattern, and a zig-zag path, as shown in Figure \ref{fig:refTraj}, superimposed over the generated `Ground truth' values for the slip parameters. The robot was positioned at $(0,0)$ and $(0,-20)$ respectively, facing in the $+x$ direction. Gaussian random noise was added to the measurements, with the random number generator seed manually set to ensure consistency between runs. The initial guess for the parameter values is $\kappa_0 = \mu_0 = 0.5$.

We implement and test the RHEC algorithm with SB-O strategy for a four-wheeled robot in simulation, comparing its performance with the FHB approach. The RHE and RHC modules are implemented in ACADO and MATLAB. For simplicity, we opt to use similar weightings as employed in \cite{Kraus2013}. The horizon length and step size used are $N_e = N_c = 15$ and $\Delta t = 0.2$ s respectively, and the steering angle is constrained to $-35^\circ \leq \delta \leq 35^\circ$.

\subsection{RHE Configuration}

Since the ACADO Code Generation tool does not yet support parameter estimation functionality, we instead augment the state and system model (\ref{eqn:sysModel}) with the slip parameter terms and two corresponding virtual controls, $u_{\kappa},u_{\mu}$, as follows: 
\begin{align}
\begin{bmatrix}
\dots, \dot{\kappa}, \dot{\mu}\end{bmatrix}^T
= 
\begin{bmatrix}
\dots, u_{\kappa}, u_{\mu}\end{bmatrix}^T.
\end{align}

In the FHB case, $u_{\kappa},u_{\mu}$ are set to zero always, fixing the parameter value over the estimation horizon. To implement SB-O, two blocking parameters $S$ and $o$ are introduced, defining the block size and overlap extent respectively. Each sampling time the bounds for $u_{\kappa},u_{\mu}$ are updated to be zero everywhere except at the start of a new block, thereby both defining the blocking structure and enforcing the desired blocking behaviour. The overlap constraint is enforced by suppressing any new blocks being started after time $t_k-o\Delta t$.

The measurement function used in these experiments is $h\left(x,u\right) = \left[x_{pos} - d\cos(\beta),y_{pos} - d\sin(\beta), \delta, u_1, u_2\right]^T$, with $d=0.4$ m being the offset of the GPS behind the rear axle centre, and we assume the same standard deviation in measurements as in \cite{Kraus2013}; $\sigma_x = \sigma_y = 0.03$ m for the GPS position measurement, $\sigma_\delta = 0.01745$ rad for the steering angle measurement, and $\sigma_{u_1} = 0.1$ m/s and $\sigma_{u_2} = 0.1$ rad/s for the actuation measurements, which are used to calculate the weighting matrix for RHE, $R_k = \text{diag}\left(\sigma_x^2,\sigma_y^2,\sigma_\delta^2, \sigma_{u_1}^2,\sigma_{u_2}^2 \right)^{-1} \in \mathbb{R}^{5\times 5}$. 

\subsection{RHC Configuration}

Let us denote the `global' space-based reference trajectory as $\Lambda_j = [x_{ref}^{(j)}, y_{ref}^{(j)}, u_{1,ref}^{(j)}, u_{2,ref}^{(j)}]^T$, which consists of a sequence of positional waypoints augmented with a desired control reference. At each sampling time $t_k$, we generate a new `local' time-based reference trajectory $\lambda$ from $\Lambda$. The local reference trajectory is defined as per Section \ref{sec:RHEC}.

This is done by taking the current position estimate and finding the closest corresponding point $(\tilde{x}_{ref}, \tilde{y}_{ref})$ along the global trajectory. Interpolation is then performed to obtain the associated control values $(\tilde{u}_{1,ref}, \tilde{u}_{2,ref})$. The next $N_c$ points along the trajectory, spaced equally $\Delta t\cdot \tilde{u}_{ref}^k$ metres along the path, are then returned as the local trajectory $\lambda_k$.

The weighting matrices for the RHC step are $V_k = \text{diag}\left(1.0, 1.0, 5.0, 5.0\right)$ and $V_T = 10 \cdot V_k$, and the steering rate constraints are $\left(-35^\circ/s \leq u_2 \leq 35^\circ/s\right)$.

\begin{figure}[tpb]
	\vspace{0.2cm}
	\centering
	\includegraphics[scale=0.6]{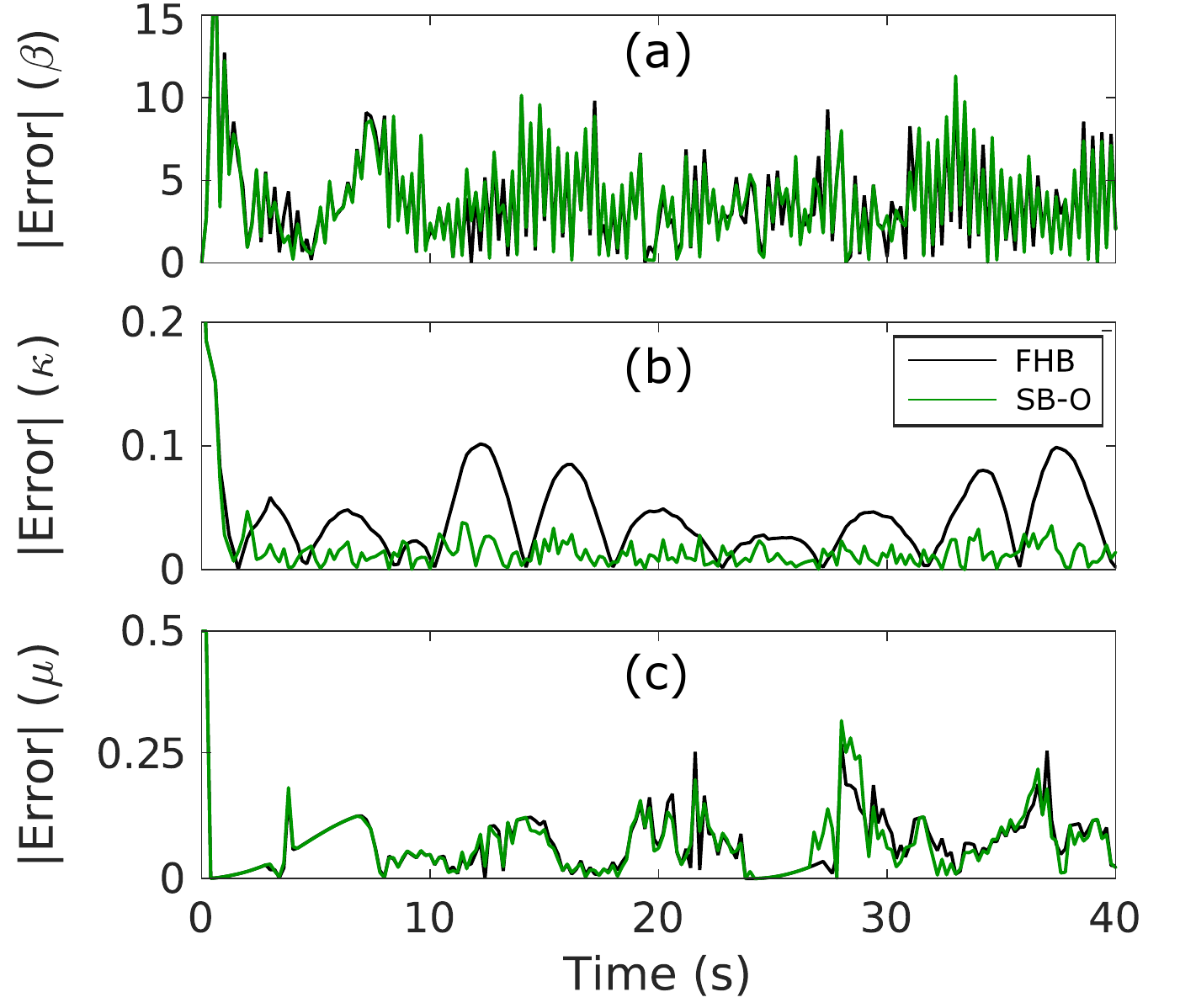} \vspace{-0.6cm}
	\caption{Comparison of estimated state and parameter errors for FHB and SB-O with $S=2$ and $o=4$ at 8 m/s. Absolute errors for (a) heading angle $\beta$ in degrees, (b) longitudinal slip $\kappa$, and (c) side slip $\mu$. SB is implemented for $\kappa$ only, and FHB is used for $\mu$.}
	\label{fig:estErrors}
	\vspace{-0.4cm}
\end{figure}

\begin{figure}[tpb]
	\centering
	\includegraphics[scale=0.53]{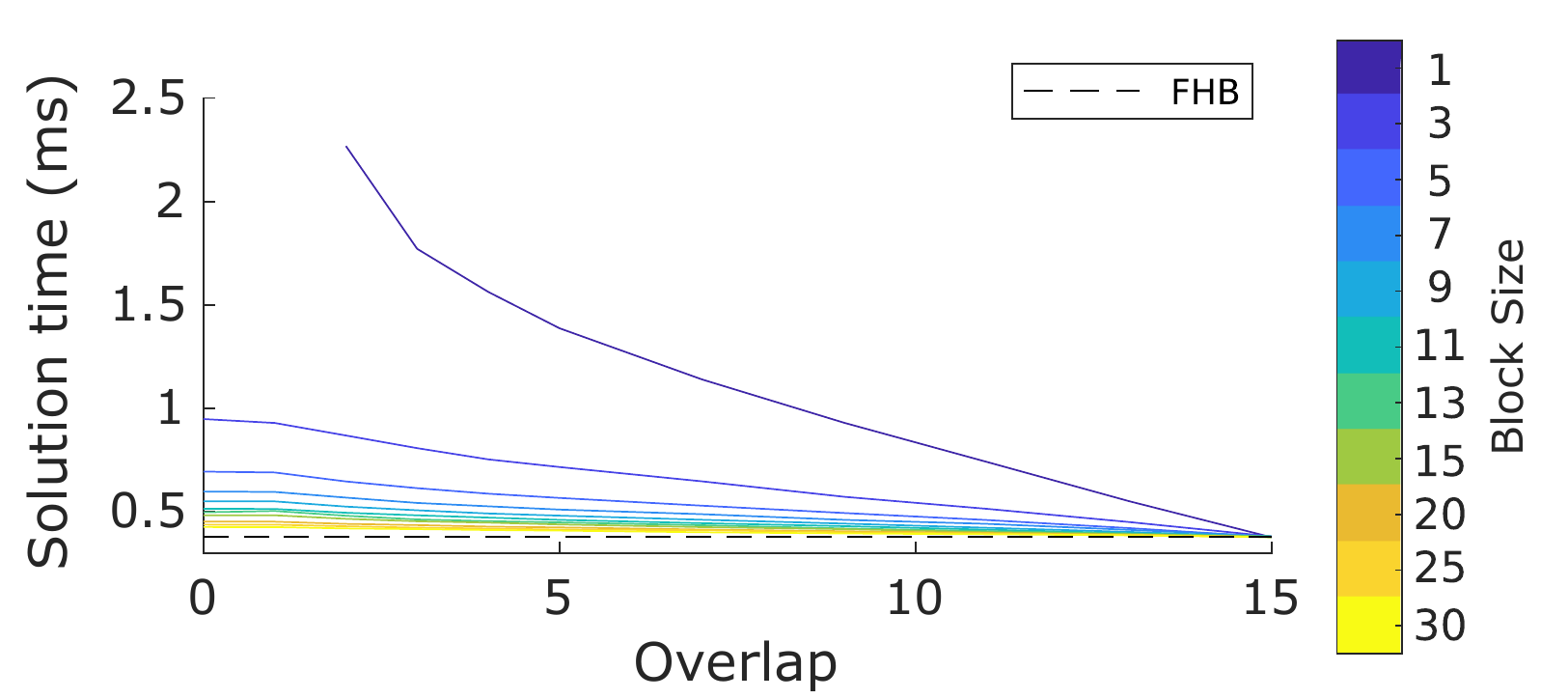} \vspace{-0.5cm}
	\caption{Comparison of computation times as function of block size $S$ and overlap extent $o$.}
	\label{fig:cpuTime}
	\vspace{-0.4cm}
\end{figure}

\section{RESULTS}

Simulations were run with the configuration as outlined in Section \ref{sec:exptSetup} above for a range of block sizes $S \in \mathscr{I}_1^{30}$ and overlap extents $o \in \mathscr{I}_0^{15}$ at each velocity $u_{1,ref} \in \mathscr{I}^{10}_{1}$ m/s. The computational effort necessary to solve the estimation problem for both the FHB and SB-O cases was also compared, as shown in Figure \ref{fig:cpuTime}, where the solution time for SB-O is seen to be competitive with the FHB method, with sub-millisecond scale solutions achieved for all except the unblocked case. While the results presented here focus primarily on higher-speed tracking, the performance of the SB-O method at lower speeds performs at least as well as the FHB method, with performance competitive with \cite{Kraus2013}\cite{Kayacan2018}.

\subsection{State and Parameter Estimation Performance}

\begin{figure}[tpb]
	\vspace{0.2cm}
	\centering
	\includegraphics[scale=0.46]{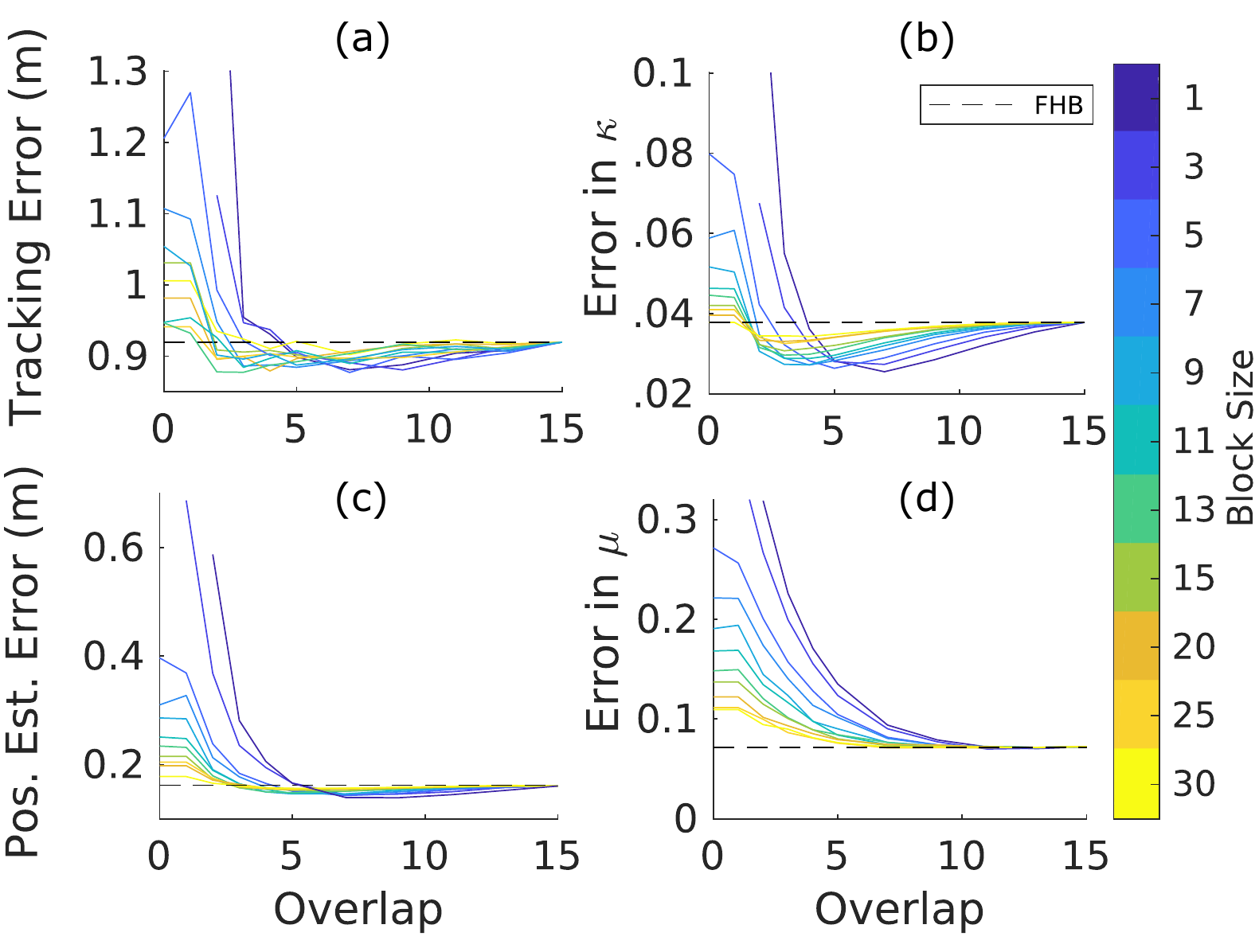}\vspace{-0.2cm}
	\caption{Performance comparison of SB-O method (applied to $\kappa$ and $\mu$) vs FHB as function of block size $S$ and overlap extent $o$ for the zig-zag trajectory at 8 m/s. Mean average error for (a) trajectory Tracking, (b) longitudinal slip $\kappa$, (c) position estimate and (d) side slip $\mu$. Note that for $S=\{1,3\}, o<2$ the estimator did not converge.}
	\label{fig:compareMAE}
\end{figure}
\begin{figure}[tpb]
	\centering
	\includegraphics[scale=0.46]{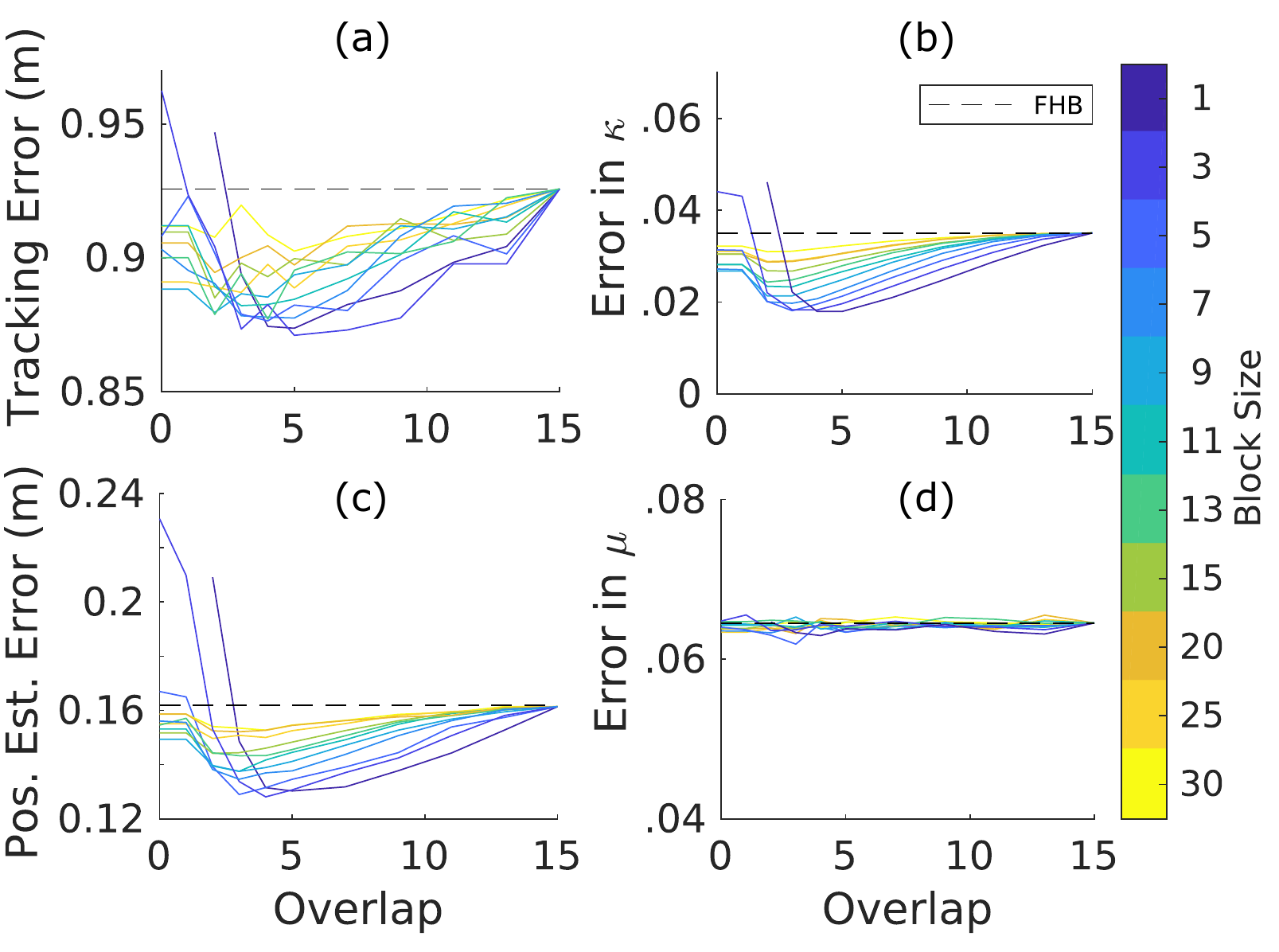}\vspace{-0.2cm}
	\caption{Performance comparison with setup as per Figure \ref{fig:compareMAE}, except SB-O is applied to $\kappa$ only ($\mu$ uses FHB in both).}
	\label{fig:compareMAE2}
	\vspace{-0.4cm}
\end{figure}
\begin{figure}[tpb]
	\vspace{0.2cm}
	\centering
	\includegraphics[scale=0.46]{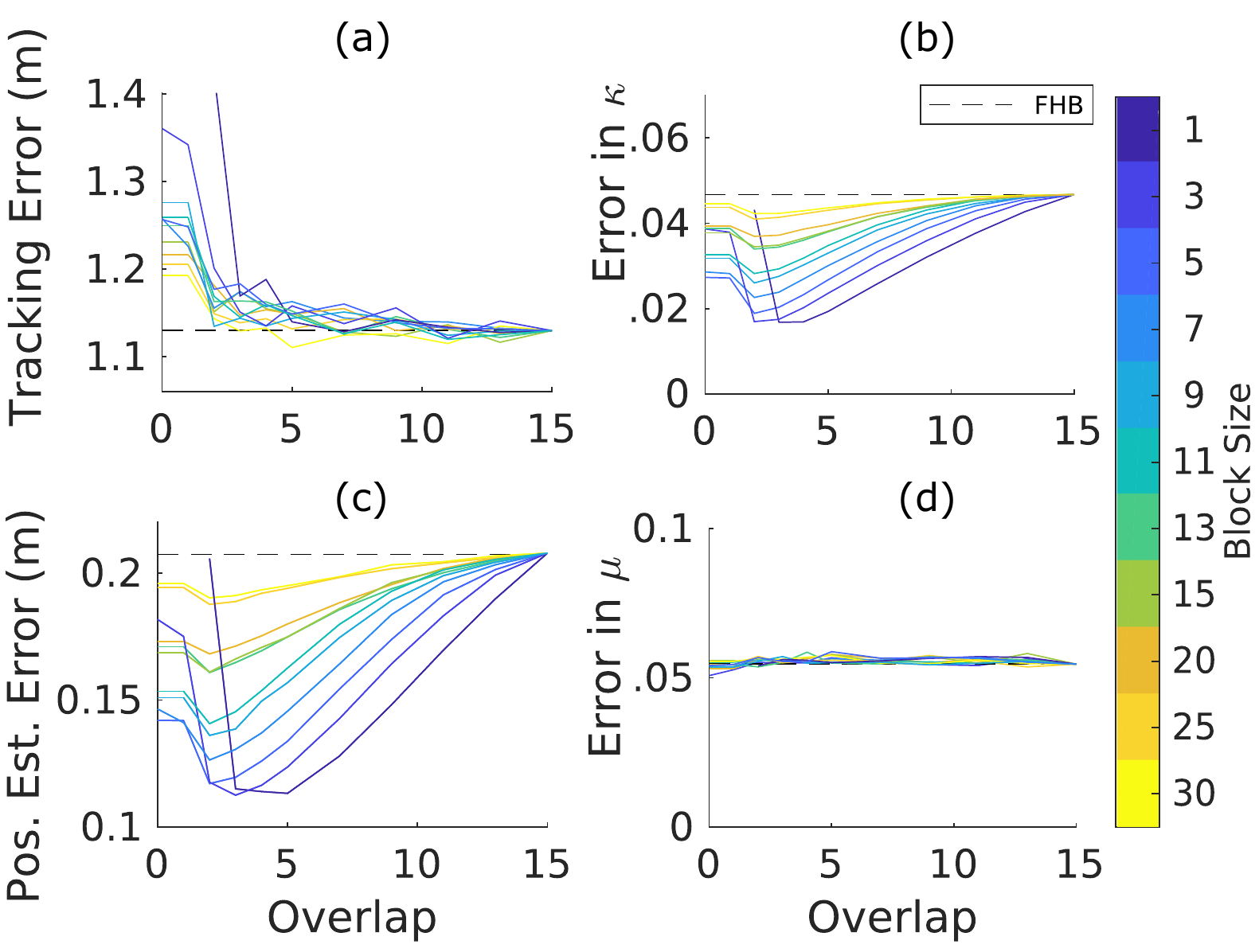}\vspace{-0.2cm}
	\caption{Performance comparison for figure-8 trajectory at 8 m/s, SB-O is applied to $\kappa$ only ($\mu$ uses FHB in both).}
	\label{fig:compareMAE3}
	\vspace{-0.4cm}
\end{figure}

The estimation performance of the SB-O and FHB methods are compared in Figure \ref{fig:estErrors} for $S=2$ and $o=4$ on the \textit{figure-8} trajectory at 8 m/s. In the comparison shown, we chose to apply SB-O to $\kappa$, but not $\mu$, as the observability of $\mu$ is proportional to the steering angle, which we found did not vary sufficiently over the $3s$ horizon for SB to yield any improvements in estimation over FHB in the tests conducted. We do see, however, that applying SB-O to $\kappa$ significantly improves the quality of the $\kappa$ estimate, while a competitive estimation quality is maintained for $\beta$ and $\mu$.

The mean average estimation errors for position and the parameter values over the full range of $(S,o)$ values tested on the \textit{zig-zag} trajectory at 8 m/s are shown in Figure \ref{fig:compareMAE} (b)-(d) for the case where SB-O is applied to both $\kappa$ and $\mu$, and in Figure \ref{fig:compareMAE2} (b)-(d) for the case where SB-O is applied to $\kappa$ while FHB is applied to $\mu$.

We observe for both cases position and $\kappa$ estimation quality improved for most choices of $(S,o)$. Increasing $o$ was seen to reduce parameter estimation errors, particularly for smaller block sizes where the periodic parameter estimation quality reduction inherent when enforcing SB would have otherwise compromised the estimate. Blocking both parameters was observed to compromise estimation quality of $\mu$ for smaller $o$, which, we believe is due to its reduced observability over the $3s$ horizon. This is supported by the results in Figure \ref{fig:compareMAE2}, where FHB is applied to $\mu$. This unambiguously improved both position and $\kappa$ estimates. Similar results are also seen for the \textit{figure-8} trajectory in Figure \ref{fig:compareMAE3}.

\subsection{Tracking Performance}

The tracking performance shown in Figures \ref{fig:compareMAE}-\ref{fig:compareMAE3} (a) indicates that smaller $S$ and $o$ correlates with increased tracking error, which is expected based on the discussions in Section \ref{sec:SB}. The tracking error was also observed to improve proportionally with estimation quality, though with some additional variance, which we expect to be due to the influence of the $\beta$ and $\delta$ estimates, which were not included in these figures. For the \textit{figure-8} case, the tracking performance was close, but slightly worse on average, despite better estimation performance, though whether this remains the case in practice will be subject to further investigation.

\section{CONCLUSION}

A RHEC framework has been implemented and tested in simulation for a 4-wheeled robot modelled with an augmented bicycle kinematics model accounting for slip. SB has been investigated as an alternative to FHB, to better capture variations in parameter values, and an overlapping-block strategy for managing periodic variations in parameter estimation quality due to the enforced blocking structure has been proposed. The performance of this strategy is assessed for a range of block-sizes $S$ and overlaps $o$, and compared to FHB methods, where it yielded improved estimation accuracy, and circumstantially better tracking performance with competitive computational efficiency.

Future work will involve experimental validation of SB-O on a field robot, and further investigation into the relationship between $S,o$ and the temporal frequency in variation of the estimated parameters. Other strategies for addressing some of the observed shortfalls in the performance of structured blocking in RHEC will also be investigated.


\begin{thebibliography}{99}

\bibitem{Fukao2000} T. Fukao, H. Nakagawa and N. Adachi, Adaptive tracking 
control of a nonholonomic mobile robot, \textit{IEEE Trans. on Robotics and Automation},
 Vol. 16, No. 5, pp. 609-615, 2000.
 
\bibitem{He2018} H. Kong and S. Sukkarieh (in press), Suboptimal Receding Horizon Estimation
via Noise Blocking, \textit{Automatica}, 2018
 
\bibitem{Kraus2013} T. Kraus, H. J. Ferreau, E. Kayacan, H. Ramon, J. De
Baerdemaeker, M. Diehl, and W. Saeys, Moving horizon estimation and
nonlinear model predictive control for autonomous agricultural vehicles, 
\textit{Computers and Electronics in Agriculture}, Vol. 98, pp. 25--33, 2013.

\bibitem{Dixon2000} W. E. Dixon, D. M. Dawson, and E. Zergeroglu, Tracking
and regulation control of a mobile robot system with kinematic disturbance:
A variable structure like approach, \textit{ASME J. Dyn. Sys., Meas., Control%
}, Vol. 122, No. 4, pp. 616-623, 2000.

\bibitem{Jiang2000} Z. P. Jiang, Robust exponential regulation of
nonholonomic systems with uncertainties, \textit{Automatica}, Vol. 36, No.
2, pp. 189-209, 2000.

\bibitem{Chen2002} Y. C. Chang and B. S. Chen, Adaptive tracking control for
nonholonomic caplygin systems, \textit{IEEE Trans. on Control Systems
	Technology}, Vol. 10, No. 1, pp. 96-104, 2002.

\bibitem{Orlando2002} M. L. Corradini, T. Leo, and G. Orlando, Experimental
testing of a discrete-time sliding mode controller for trajectory tracking
of a wheeled mobile robot in the presence of skidding effects, \textit{%
	Journal of Robotic Systems}, Vol. 19, No. 4, pp. 177-188, 2002.

\bibitem{Wang2008TRO} D. Wang and C. B. Low, Modeling and analysis of
skidding and slipping in wheeled mobile robots: control design perspective, 
\textit{IEEE Trans. on Robotics}, Vol. 24, No. 3, pp. 676-687, 2008.

\bibitem{Wang2010} C. B. Low and D. Wang, Maneuverability and path following
control of wheeled mobile robot in the presence of wheel skidding and
slipping, \textit{Journal of Field Robotics}, Vol. 27, No. 2, pp. 127-144,
2010.

\bibitem{Savkin2013} A. S. Matveev, M. Hoy, J. Katupitiya, and A. V. Savkin,
Nonlinear sliding mode control of an unmanned agricultural tractor in the
presence of sliding and control saturation, \textit{Robotics and Autonomous
	Systems}, Vol. 61, No. 9, pp. 973-987, 2013.

\bibitem{Ramon2015} E. Kayacan, E. Kayacan, H. Ramon, and Wouter Saeys,
Robust tube-Based decentralized nonlinear model predictive control of an
autonomous tractor-trailer system, \textit{IEEE/ASME Trans. on Mechatronics}%
, Vol. 20, No. 1, pp. 447-456, 2015.

\bibitem{Alamo2011} R. Gonz\'{a}lez, M. Fiacchini, J. L Guzm\'{a}n, T. \'{A}%
lamo, and F. Rodr\'{\i}guez, Robust tube-based predictive control for mobile
robots in off-road conditions, \textit{Robotics and Autonomous Systems},
Vol. 59, No. 10, pp. 711-726, 2011.

\bibitem{Yi2009} J. Yi, H. Wang, J. Zhang, D. Song, S. Jayasuriya and J. Liu, 
Kinematic modeling and analysis of skid-steered mobile robots with applications 
to low-cost inertial-measurement-unit-based motion estimation, \textit{%
IEEE Trans. on Robotics}, Vol. 25, No. 5, pp. 1087-1097, 2009.

\bibitem{Sukkarieh2001} G. Dissanayake, S. Sukkarieh, E. Nebot, and H.
Durrant-Whyte, The aiding of a low-cost strapdown inertial measurement unit
using vehicle model constraints for land vehicle applications, \textit{IEEE
	Trans. on Robotics and Automation}, Vol. 17, No. 5, pp. 731-747, 2001.

\bibitem{Backman2012} J. Backman, T. Oksanen, A. Visala, Navigation system
for agricultural machines: nonlinear model predictive path tracking, \textit{%
Computers and Electronics in Agriculture}, Vol. 82, pp. 32-43, 2012.

\bibitem{Kayacan2014} E. Kayacan, E. Kayacan, H. Ramon, and Wouter Saeys,
Distributed nonlinear model predictive control of an autonomous
tractor--trailer system, \textit{Mechatronics}, Vol. 24, No. 8, pp. 926-933,
2014.

\bibitem{Kayacan2018} E. Kayacan, S. N. Young, J. M. Peschel, and G. Chowdhary,
High‐precision control of tracked field robots in the presence of unknown traction 
coefficients, \textit{Journal of Field Robotics}, pp. 1-13, 2018.

\bibitem{Thuilot2006} R. Lenain, B. Thuilot, C. Cariou, and P. Martinet,
High accuracy path tracking for vehicles in presence of sliding: application
to farm vehicle automatic guidance for agricultural tasks, \textit{%
Autonomous Robots}, Vol. 21, No. 1, pp 79-97, 2006.

\bibitem{Thuilot2010} R. Lenain, B. Thuilot, C. Cariou, and P. Martinet,
Mixed kinematic and dynamic sideslip angle observer for accurate control of
fast off-road mobile robots, \textit{Journal of Field Robotics}, Vol. 27,
No. 2, pp. 181-196, 2010.

\bibitem{Angelova2007} A. Angelova, L. Matthies, D. Helmick, and P. Perona,
Learning and prediction of slip from visual information, \textit{Journal of
Field Robotics}, Vol. 24, No. 3, pp. 205-231, 2007.

\bibitem{Collier2016} C. J. Ostafew, A. P. Schoellig, T. D. Barfoot, and J.
Collier, Learning-based nonlinear model predictive control to improve
vision-based mobile robot path tracking, \textit{Journal of Field Robotics},
Vol. 33, No. 1, pp. 133-152, 2016.

\bibitem{Ostafew2016} C. J. Ostafew, A. P. Schoellig, and T. D. Barfoot,
Robust constrained Learning-based NMPC enabling reliable mobile robot path
tracking, \textit{The Int. Journal of Robotics Research}, Vol. 35, No. 13,
pp. 1547-1563, 2016.

\bibitem{Aswani2013} A. Aswani, H. Gonzalez, S. Shankar Sastry and C. Tomlin, 
Provably safe and robust learning-based model predictive control, 
\textit{Automatica}, Vol. 49, pp. 1216-1226, 2013.

\bibitem{Abbeel2006} P. Abbeel, M. Quigley, and A. Ng, 
Using inaccurate models in reinforcement learning, 
\textit{In Proceedings of the 23rd International Conference 
	on Machine Learning}, pp. 1-8, ACM, 2006.

\bibitem{LaValle2006} S. M. LaValle, Planning Algorithms, \textit{Cambridge University Press}, pp. 722--726, 2006.

\bibitem{Snider2009} J. M. Snider, Automatic steering methods for autonomous 
automobile path tracking, \textit{Robotics Institute, Pittsburgh, PA, Tech. Rep. CMU-RITR-09-08}, 2009.

\bibitem{Kong2015} J. Kong, M. Pfeiffer, G. Schildbach and F. Borrelli, 
Kinematic and dynamic vehicle models for autonomous driving control design, 
\textit{2015 IEEE Intelligent Vehicles Symposium (IV)}, pp. 1094--1099, Seoul, South Korea, 2015.

\bibitem{Ferreau2012} H. Ferreau, T. Kraus, M. Vukov, W. Saeys and M. Diehl, 
High-speed moving horizon estimation based on automatic code generation, 
\textit{2012 IEEE 51st Annual Conference on Decision and Control (CDC)}, 
pp. 687--692, Maui, HI, 2012.

\bibitem{Rawlings2001} C. V. Rao, J. B. Rawlings, and J. H. Lee, Constrained
linear state estimation--a moving horizon approach, \textit{Automatica},
Vol. 37, No. 10, pp. 1619-1628, 2001.

\bibitem{Voelker2013} A. Voelker, K. Kouramas, and E. Pistikopoulos, Moving
horizon estimation: error dynamics and bounding error sets for robust
control, \textit{Automatica}, Vol. 49, No. 4, pp. 943--948, 2013.

\bibitem{Pannocchia2015} B. Morabito, M. Kogel, E. Bullinger, G. Pannocchia,
and R. Findeisen, Simple and efficient moving horizon estimation based on
the fast gradient method, \textit{Proc. of 5th IFAC Conference on NMPC}, pp.
428-433, Seville, Spain, 2015.

\bibitem{Lee2015} T. Y. Jung, H. Jang, and J. H. Lee, Move blocking strategy
applied to re-entrant manufacturing line scheduling, \textit{Int. Journal of
	Control, Automation and Systems}, Vol. 13, No. 2, pp. 410-418, 2015.

\bibitem{Shekhar2015} R. Shekhar and C. Manzie, Optimal move blocking strategies 
for model predictive control, \textit{Automatica}, Vol. 61, pp. 27--34, 2015.

\bibitem{Wang2001} L. Wang, Continuous time model predictive control design
using orthonormal functions, \textit{Int. Journal of Control}, Vol. 74, No.
16, pp. 1588-1600, 2001.

\bibitem{Diehl2002} M. Diehl, H. G. Bock, J. P. Schlöder, R. Findeisen, Z. Nagy and 
F. Allgöwer, Real-time optimization and nonlinear model predictive control of 
processes governed by differential-algebraic equations, \textit{%
	Journal of Process Control}, Vol. 12, No. 4, pp. 577--585, 2002.

\bibitem{Houska2011} B. Houska, H. J. Ferreau and M. Diehl, {ACADO} {T}oolkit -- 
{A}n {O}pen {S}ource {F}ramework for {A}utomatic {C}ontrol and {D}ynamic {O}ptimization, 
\textit{Optimal Control Applications and Methods}, Vol. 32, No. 3, pp. 298--312, 2011.

\bibitem{Ferreau2014} H. J. Ferreau, C. Kirches, A. Potschka, H. G. Bock and M. Diehl, 
{qpOASES}: A parametric active-set algorithm for quadratic programming, 
\textit{Mathematical Programming Computation}, Vol. 6, No. 4, pp. 327--363, 2014.
\end{thebibliography}
\end{document}